\documentclass[a4paper,12pt]{article}
\usepackage{latexsym,amssymb,enumerate,amsmath,epsfig,amsthm,authblk,dsfont,bm}
\usepackage[margin=1in]{geometry}
\usepackage{setspace,color}
\usepackage{tikz}
\usepackage{floatrow}
\usepackage{graphicx,subfig}
\usepackage[ruled]{algorithm2e}
\usepackage{epstopdf}
\usepackage{grffile}
\usepackage{makecell}
\usepackage{graphbox} 
\usepackage{lineno}
\usepackage[numbers,sort]{natbib}

\usepackage{hyperref}
\hypersetup{
	colorlinks=true,
	linkcolor=blue,
	citecolor=blue,
	filecolor=blue,      
	urlcolor=blue
}

\newcommand{\bb}{\mathbf{b}}

\newcommand{\RR}{\mathds{R}}

\newcommand{\bx}{\mathbf{x}}

\newcommand{\bn}{\mathbf{n}}

\newcommand{\cN}{\mathcal{N}}
\newcommand{\cL}{\mathcal{L}}

\newcommand{\Per}{\mathrm{Per}}

\newcommand{\red}[1]{{#1}}

\begin{document}
\title{Connections between Operator-splitting Methods and Deep Neural Networks with Applications in Image Segmentation }
\date{\textit{In memory of Prof. Zhongci Shi}}
\author{
	Hao Liu\thanks{Department of Mathematics, Hong Kong Baptist University, Kowloon Tong, Hong Kong. Email: haoliu@hkbu.edu.hk. The work of Hao Liu is partially supported by HKBU 179356, NSFC 12201530 and HKRGC ECS 22302123.} , 
	Xue-Cheng Tai\thanks{Norwegian Research Centre (NORCE), Nyg{\aa}rdsgaten 112, 5008 Bergen, Norway.  Email: xtai@norceresearch.no, xuechengtai@gmail.com. The work of Xue-Cheng Tai is partially supported by NSFC/RGC grant N-HKBU214-19 and NORCE Kompetanseoppbygging program.},
	 Raymond Chan \thanks{Department of Mathematics, City University of Hong Kong and Hong Kong Center for Cerebro-Cardiovascular Health Engineering, Hong Kong SAR. raymond.chan@cityu.edu.hk. The work of Raymond Chan is partially supported by
HKRGC GRF grants CityU1101120, CityU11309922, CRF grant C1013-21GF, and HKRGC-NSFC Grant N\_CityU214/19.}

}
\maketitle

% \begin{center}

% \end{center}

\begin{abstract}
	Deep neural network is a powerful tool for many tasks. Understanding why it is so successful and providing a mathematical explanation is an important problem and has been one popular research direction in past years. In the literature of mathematical analysis of deep neural networks, a lot of works is dedicated to establishing representation theories. How to make connections between deep neural networks and mathematical algorithms is still under development. In this paper, we give an algorithmic explanation for deep neural networks, especially in their connections with operator splitting. We show that with certain splitting strategies, operator-splitting methods have the same structure as networks. Utilizing this connection and the Potts model for image segmentation, two networks inspired by operator-splitting methods are proposed. The two networks are essentially two operator-splitting algorithms solving the Potts model. Numerical experiments are presented to demonstrate the effectiveness of the proposed networks.
\end{abstract}
\section{Introduction}
In past decades, deep neural network has emerged as a very successful technique for various fields. It has demonstrated impressive performances in many tasks, such as image processing, object detection, and natural language processing. In some tasks, deep neural networks even outperform humans. 

Due to great successes of neural networks, over the past several years, a lot of works has been devoted to the mathematical understanding of neural networks and explaining their success. Representation theories for function learning are studied in \cite{barron1993universal,yarotsky2017error,chen2019efficient,lu2021deep,liu2023deep} for feedforward neural networks, in \cite{zhou2020theory,zhou2020universality} for convolutional neural networks, and in \cite{oono2019approximation,liu2022benefits,liu2021besov} for convolutional residual networks. Recently, theoretical results for learning operators are developed in \cite{liu2022deepop,lanthaler2022error,bhattacharya2021model}. The works mentioned above show that as long as the network depth and width are sufficiently large, deep neural networks can approximate any function or operator within a certain class to arbitrary accuracy. These works focus on the existence of a good approximator with desired approximation error and use techniques from approximation theory. Recently, analysis of neural ordinary differential equations from the perspective of optimal control was conducted in \cite{ruiz2023neural}.
In this paper, we investigate the power of neural networks from another perspective: the network structure. We will give an algorithmic explanation of neural networks, especially in their connections with operator-splitting algorithms.

The operator-splitting method is a class of powerful methods for numerically solving complicated problems. The general idea is to decompose a difficult problem into several subproblems which will be solved sequentially or simultaneously. Operator-splitting methods have been widely used on solving partial differential equations \cite{glowinski2019finite,glowinski2020numerical, lu1992parallel,marchuk1990splitting,wang2022efficient}, image processing \cite{deng2019new,duan2022fast,liu2022operator, liu2021color}, surface reconstruction \cite{he2020curvature}, obstacle problem \cite{liu2023fast}, inverse problem \cite{glowinski2015penalization}, and computational fluid dynamics \cite{bonito2017operator,mrad2022splitting}, etc. We refer readers to \cite{macnamara2016operator, glowinski2016some} for some survey discussions.

Image segmentation is an important subject in many fields, such as medical imaging and object detection. Many mathematical models and algorithms have been developed for image segmentation \cite{chan1999active,kass1988snakes, caselles1997geodesic, mumford1989optimal,wang2017efficient, chan2018convex, cai2017three, chan2014two}. 
In \cite{bae2011global,yuan2010study}, the segmentation problem is formulated as a min cut or max flow problem. One important model for image segmentation is the Potts model, which was first proposed for statistical mechanics \cite{potts1952some}. In fact, the well-known Chan-Vese model \cite{chan1999active} is a special case of the Potts model.  In \cite{wei2015primal}, detailed explanations are given to show that  the Potts model is equivalent to a continuous min cut and max flow problem \cite{wei2015primal}. Efficient algorithms for the Potts model are studied in \cite{sun2021efficient,yuan2010study}. We suggest readers to survey \cite{tai2021potts} for a comprehensive discussion on the Potts model. Recently, many deep learning methods for image segmentation are also proposed \cite{ronneberger2015u,zhou2018unet++,fan2020ma}.

Following \cite{tai2023prep} and \cite{liu2023prep}, we focus on the building block of neural networks and operator-splitting methods and make connections between them. We first introduce the structure of a standard neural network in this work. 
Then we discuss popular operator-splitting strategies: sequential splitting and parallel splitting. We show that for certain splitting strategies, the resulting splitting algorithm is equivalent to a neural network, whose depth and width are determined by the splitting strategy. Such a connection is also observed and utilized in \cite{lan2022dosnet}, which is used to solve nonlinear partial differential equations by neural networks. We will apply this connection to the Potts model for image segmentation, and propose two networks inspired by operator-splitting methods. As the proposed networks are derived from the Potts model, they contain explicit regularizers that have physical meaning. The effectiveness of the proposed networks are demonstrated by numerical experiments.

This paper is structured as follows: We briefly introduce deep neural networks and operator-splitting methods in Section \ref{sec.NN} and \ref{sec.splitting}, respectively. The Potts model for image segmentation is discussed in Section \ref{sec.potts}. We present the two networks inspired by operator-splitting methods in Section \ref{sec.model1} and \ref{sec.model2}, respectively. This paper is concluded in Section \ref{sec.conclusion}.

\section{Deep neural networks}
\label{sec.NN}
Deep neural networks have been successfully applied to many problems and have achieved remarkable performances. There are many networks architectures, such as feedforward neural networks (FNN) \cite{cybenko1989approximation}, convolutional neural networks (CNN) \cite{lecun1989backpropagation}  and residual neural networks (ResNet) \cite{he2016deep}. It is easy to show that any CNN and ResNet can be realized by FNN \cite{zhou2020theory,zhou2020universality,oono2019approximation,liu2021besov}. In this paper, we focus on FNN and show their connections with operator-splitting methods.

The building blocks of FNNs are layers. An FNN is a composition of multiple layers, each of which takes an input $\bx\in \RR^{d}$ and outputs a vector in $\RR^{d'}$ for some integers $d,d'>0$. Each layer has the form of
\begin{align}
	\cL(W,\bb,\sigma;\bx)=\sigma(W\bx+\bb),
	\label{eq.layer}
\end{align}
 where $W\in \RR^{d\times d'}$ is a weight matrix, $\bb\in \RR^{d'}$ is a bias term, and $\sigma(\cdot)$ is the activation function. Popular choices of $\sigma$ include the rectified linear unit (ReLU), sigmoid function and tanh function. The computation of a layer consists of two parts, a linear part $W\bx+\bb$ and a nonlinear part $\sigma(\cdot)$. These two parts are conducted sequentially. Later we will show the similarity between this structure and operator splitting methods.

An FNN with depth $L$ is a composition of $L-1$ layers followed by a fully connected layer:
\begin{align}
	f(\bx)=W_L\cL_L(W_{L-1},\bb_{L-1},\sigma_{L-1})\circ \cdots \circ \cL_1(W_1,\bb_1,\sigma_1;\bx)+\bb_L,
\end{align}
where $W_l\in \RR^{d_{l-1} \times d_l}$, $\bb_l\in \RR^{d_l}$ and $\sigma_l$ denote the weight matrix, bias and activation function in the $l$-th layer with $d_0=d$ being the input dimension and $d_L$ being the output dimension. We call $\max_l d_l$ the width of $f$. The computation of $f(\bx)$ can be taken as passing the input $\bx$ to $L$ layers sequentially. 

\section{Operator-splitting methods}
\label{sec.splitting}
The operator-splitting method is a powerful method for solving complicated problems, such as time evolution problems and optimization problems. The general idea is to decompose a complicated problem into several easy-to-solve subproblems, so that each subproblem can be solved explicitly or efficiently. The first operator-splitting method according to \cite{chorin1978product} is the famous Lie scheme introduced in \cite{lie1970theorie} to solve the initial problem from the dynamical system:
\begin{align}
	\begin{cases}
		{\displaystyle \frac{dX}{dt}}+(A+B)X=0 \mbox{ in } (0,T],\\
		X(0)=X_0,
	\end{cases}
	\label{eq.Lie.ivp}
\end{align}
where $X_0\in \RR^d$ for some integer $d>0$, and $A,B\in \RR^{d\times d}$. In the Lie scheme, one solves (\ref{eq.Lie.ivp}) by decomposing $A$ and $B$ into two subproblems. In each subproblem, $X$ is governed by one operator and evolves for a small time step. One can show that this scheme is first-order accurate in time. Later, a second order splitting scheme, the Strang scheme, was introduced in \cite{strang1968construction}. We refer this type of splitting scheme as sequential scheme as the subproblems are solved sequentially. Another type of splitting strategy is parallel splitting \cite{lu1992parallel}, which solves the subproblems simultaneously and then takes the average. Such a method is proved to be first-order accurate in time. Due to the simplicity of the ideas and versatility, operator-splitting methods have been widely used in solving partial differential equations \cite{glowinski2019finite,bonito2017operator,bukavc2013fluid,lu1992parallel,marchuk1990splitting} in which the operators in the PDE are decomposed into subproblems.

The operator-splitting method is also a popular choice to solve optimization problems. Consider an optimization problem in which one needs to minimize a functional. One may first derive the optimality condition of the minimizer and associate it with an initial value problem in the flavor of gradient flow. Then solving the optimization problem is converted into finding the steady-state solution of the initial value problem. The initial value problem can be solved by operator-splitting methods. Such a strategy has been used in \cite{deng2019new,liu2021color,liu2023elastica,duan2022fast,he2020curvature}. For optimization problems, the alternating direction method of multipliers (ADMM) has been extensively studied in past decades \cite{sun2021efficient,yuan2010study,bae2017augmented,yashtini2016fast}. Indeed, ADMM is a special type of operator-splitting methods. 
We suggest readers to \cite{glowinski2016some,glowinski2017splitting} for a comprehensive discussion on operator-splitting methods.

In the rest of this section, we focus on the following initial value problem
\begin{align}
	\begin{cases}
		\displaystyle \frac{\partial u}{\partial t}=\sum_{k=1}^K(A_ku + B_k(u)+g_k) \mbox{ on } \Omega\times (0,T],\\
		u(0)=u_0,
	\end{cases}
	\label{eq.ivp}
\end{align}
where $K>0$ is a positive integer, $\Omega$ is our computational domain, $A_ku$ is a linear operator of $u$, $B_k(u)$ is an operator on $u$ that might be nonlinear,  and $g_k$'s are functions defined on $\Omega$. We will discuss sequential splitting and parallel splitting schemes for solving (\ref{eq.ivp}), and show their connections to deep neural networks. In the following, assume the time domain is discretized into $N$ subintervals. Denote $\Delta t=T/N$ and $t^n=n\Delta t$. We use $u^n$ to denote our numerical solution at $t^n$.

\subsection{Sequential splitting}
For sequential splitting, a simple one is the Lie scheme. Given $u^n$, we can update $u^{n+1}$ by $K$ substeps:

\noindent For $k=1,...,K$, solve
\begin{align}
	\begin{cases}
		{\displaystyle \frac{\partial v}{\partial t}}= A_kv + B_k(v)+g_k \mbox{ on } \Omega\times (t^n,t^{n+1}],\\
		v(t^n)=u^{n+\frac{k-1}{K}},
	\end{cases}
\label{eq.sequential}
\end{align}
and set $u^{n+\frac{k}{K}}=v(t^{n+1})$. One can show that when $B_k$'s are linear operators and (\ref{eq.sequential}) is time-discretized by the forward Euler method, scheme (\ref{eq.sequential}) is first-order accurate in time.

Equation (\ref{eq.sequential}) is the building block for this scheme. Note that to solve (\ref{eq.sequential}), we can further apply a Lie scheme. Starting from $u^{n+\frac{k-1}{K}}$, we update $u^{n+\frac{k}{K}}$ by two parts $u^{n+\frac{k-1}{K}}\rightarrow \bar{u}^{n+\frac{k}{K}} \rightarrow u^{n+\frac{k}{K}}$:

\noindent Part 1: Solve
\begin{align}
	\begin{cases}
		{\displaystyle \frac{\partial v}{\partial t}}= A_kv +g_k \mbox{ on } \Omega\times (t^n,t^{n+1}],\\
		v(t^n)=u^{n+\frac{k-1}{K}},
	\end{cases}
	\label{eq.sequential.step1}
\end{align}
and set $\bar{u}^{n+\frac{k}{K}}=v(t^{n+1})$.  %??? 

\noindent Part 2: Solve
\begin{align}
	\begin{cases}
		\displaystyle \frac{\partial v}{\partial t}=B_k(v) \mbox{ on } \Omega\times (t^n,t^{n+1}],\\
		v(t^n)=\bar{u}^{n+\frac{k}{K}},
	\end{cases}
	\label{eq.sequential.step2}
\end{align}
and set $u^{n+\frac{k}{K}}=v(t^{n+1})$.

By time-discretizing (\ref{eq.sequential.step1}) by the forward Euler method, we get
\begin{align}
	\frac{\bar{u}^{n+\frac{k}{K}}-u^{n+\frac{k-1}{K}}}{\Delta t}=A_ku^{n+\frac{k-1}{K}} +g_k,
\end{align}
implying that
\begin{align}
	\bar{u}^{n+\frac{k}{K}}&=u^{n+\frac{k-1}{K}}+ \Delta t  A_ku^{n+\frac{k-1}{K}}+\Delta t g_k \nonumber\\
	&=(I+\Delta t A_k)u^{n+\frac{k-1}{K}}+\Delta t g_k,
\end{align}
where $I$ denotes the identity operator: $Iu=u$.

By time-discretizing (\ref{eq.sequential.step2}) using the backward Euler method, we get
\begin{align}
	\frac{u^{n+\frac{k}{K}}-\bar{u}^{n+\frac{k}{K}}}{\Delta t}=B_k(u^{n+\frac{k}{K}}).
\end{align}
Denoting the resolvant operator for $u^{n+\frac{k}{K}}$ by $\rho_k$, i,e. $\rho_k = (I -\Delta t  B_k)^{-1}$,  we have
\begin{align}
	u^{n+\frac{k}{K}}=\rho_k(\bar{u}^{n+\frac{k}{K}})= \rho_k((I+\Delta t A_k)u^{n+\frac{k-1}{K}}+\Delta t g_k).
	\label{eq.sequential.block}
\end{align}

\paragraph{Connections to FNN.} Comparing (\ref{eq.sequential.block}) and (\ref{eq.layer}), we see that they have the same structure. By choosing,
\begin{align}
	L=K+1, W_k=(I+\Delta t A_k), \ \bb_k=\Delta t g_k, \ \sigma_k=\rho_k
\end{align}
and $W_{K+1}=I, \ \bb_{K+1}=0$, the sequential splitting scheme is an FNN with $K+1$ layers, where the $k$-th substep corresponds to the $k$-th layer. In other words, any FNN with $W_{K+1}=I, \ \bb_{K+1}=0$ and activation functions $\rho_k$'s is a sequential splitting scheme solving some initial value problem.
\subsection{Parallel splitting}
Using parallel splitting \cite{lu1992parallel}, we solve (\ref{eq.ivp}) by first solving $K$ parallel substeps:

\noindent For $k=1,...,K$, solve
\begin{align}
	\begin{cases}
		{\displaystyle \frac{\partial v}{\partial t}}= KA_kv +KB_k(v)+Kg_k \mbox{ on } \Omega\times (t^n,t^{n+1}],\\
		v(t^n)=u^{n}
	\end{cases}
\label{eq.parallel}
\end{align}
and set $u^{n+1,k}=v(t^{n+1})$. Then compute
\begin{align}
	u^{n+1}=\frac{1}{K}\sum_{k=1}^K u^{n+1,k}.
\end{align}
One can show that when $B_k$'s are linear operators and (\ref{eq.parallel}) is solved by the forward Euler method, the parallel splitting scheme is first-order accurate in time.
Note that the number of operators $K$ is multiplied into the right hand side of (\ref{eq.parallel}).

Similar to (\ref{eq.sequential.step1})--(\ref{eq.sequential.step2}), we can use the same idea to solve (\ref{eq.parallel}). After some derivations, we get
\begin{align}
	u^{n+1,k}=\rho_k\left((I+\Delta t KA_k)u^{n}+\Delta t Kg_k\right)
	\label{eq.parallel.block}
\end{align}
and 
\begin{align}
	u^{n+1}=\frac{1}{K}\sum_{k=1}^K\rho_k\left((I+\Delta t KA_k)u^{n}+\Delta t Kg_k\right).
\end{align}

\paragraph{Connections to FNN.} Again, (\ref{eq.parallel.block}) and (\ref{eq.layer}) have the same structure. Set $L=2$,
\begin{align}
	W_1=\begin{bmatrix}
		A_1 &0 & \cdots & 0 \\
		0 & A_2 & \cdots & 0\\
		\vdots & \cdots &\ddots & \vdots\\
		0 & \cdots & \cdots &A_K 
	\end{bmatrix}, \ \bb_1=\begin{bmatrix}
	g_1\\ \vdots \\ g_K
\end{bmatrix} \ \sigma_1=\begin{bmatrix}
\rho_1\\ \vdots \\ \rho_K
\end{bmatrix}, \ W_2=\frac{1}{K}I, \ \bb_2=0,
\label{eq.parallel.weight}
\end{align}
where $\sigma_1$ is applied elementwise: 
\begin{align}
	\sigma_1\left( \begin{bmatrix}
		u_1 \\ \vdots \\ u_K
	\end{bmatrix}\right)= \begin{bmatrix}
	\rho_1(u_1) \\ \vdots \\ \rho_K(u_K)
\end{bmatrix},
\end{align}
then the parallel splitting scheme is an FNN with 2 layers. In other words, any 2-layer FNN with $\sigma_1,W_2,\bb_2$ given in the format of (\ref{eq.parallel.weight}) is a parallel splitting scheme solving some initial value problems.

\section{Potts model and image segmentation}\label{sec.potts}
We next focus on image segmentation and utilize the relations between operator-splitting methods and FNNs to derive new networks. 

We start with the Potts model which has close relations to a large class of image segmentation models. Let $\Omega$ be the image domain. The continuous two-phase Potts model is given as \cite{chambolle2011first, boykov2006graph,boykov2004experimental,yuan2010continuous,Bae2014a}
\begin{align}
	\begin{cases}
		\min\limits_{\Sigma_0,\Sigma_1} \left\{\displaystyle \sum_{k=0}^1 \displaystyle\int_{\Sigma_k} f_k(\bx) d\bx+\frac{1}{2} \displaystyle\sum_{k=0}^1 \Per(\Sigma_k) \right\},\\
		\Sigma_0\cup \Sigma_1=\Omega,\qquad
		\Sigma_0\cap \Sigma_1=\emptyset,
	\end{cases}
\label{eq.potts0}
\end{align}
where $\Per(\Sigma_k)$ is the perimeter of $\Sigma_k$, and $f_k$'s are nonnegative weight functions. In (\ref{eq.potts0}), the first term is a data fidelity term which depends on the input image $f$. The second term is a regularization term which penalizes the perimeter of the segmented region.

A popular choice of $f_k$ is 
\begin{align*}
	f_k(\bx)= (f(\bx)-c_k)^2,
\end{align*}
where $c_k$ is the estimated mean density of $f(\bx)$ on $\Sigma_k^*$. Here we use $\Sigma_k^*$ to denote the `optimal' segmentation region. With this choice and the use of a binary function $v$ to represent the foreground of the segmentation results, we rewrite the two-phase Potts model as
\begin{align}
	\min\limits_{v\in\{0,1\}} \left\{ \displaystyle\int_{\Omega} (f(\bx)-c_0)^2v+ (f(\bx)-c_1)^2(1-v) d\bx+ \lambda \int_{\Omega} |\nabla v|d\bx \right\}
	\label{eq.potts1}
\end{align}
for some constant $\lambda>0$. Model (\ref{eq.potts1}) is the well-known Chan-Vese model \cite{chan2001active}.

The first integral in (\ref{eq.potts1}) is linear in $v$. As one can take $c_k$ and $\Sigma_k^*$ as functions depending on $f$, in this case, $f_k$'s in (\ref{eq.potts0}) are functions of the input image $f$ only.
We can thus rewrite the Potts model as 
\begin{align}
	\min_{v\in\{0,1\}}\int_{\Omega} F(f)vd\bx +\lambda \int_{\Omega}|\nabla v| d\bx,
	\label{eq.potts}
\end{align}
where $F(f)$ is a region force depending on the input image $f$. In the following sections, we will discuss two relaxations of the Potts model (\ref{eq.potts}) and correspondingly propose two new networks for image segmentation.

\section{Operator-splitting method inspired networks for image segmentation: Model I}
\label{sec.model1}

Our first model utilizes (\ref{eq.potts}) and the double-well potential and follows \cite{liu2023prep}. 

\subsection{Model formulation}
Model (\ref{eq.potts}) requires the function $v$ to be binary. One relaxation of (\ref{eq.potts}) using the double-well potential is 
\begin{align}
	\min_v \int_{\Omega} F(f)vd\bx +\lambda \cL_{\varepsilon}(v) d\bx,
	\label{eq.potts.relax}
\end{align}
with 
\begin{align*}
	\cL_{\varepsilon}(v)=\int_{\Omega} \left[ \frac{\varepsilon}{2}|\nabla v|^2 + \frac{1}{\varepsilon} v^2(1-v)^2\right] d\bx.
\end{align*}
It is shown in \cite{modica1977esempio,modica1987gradient} that $\cL_{\varepsilon}(v)$ converges to $C\Per(\Sigma_1)$  in the sense of $\Gamma$-convergence as $\varepsilon\rightarrow 0$ for some constant $C$.

Denote the minimizer of (\ref{eq.potts.relax}) by $u$. It satisfies the optimality condition
\begin{align}
	F(f)-\lambda\varepsilon\nabla^2u +\frac{2\lambda}{\varepsilon}(2u^3-3u^2+u)=0 \mbox{ in }  \Omega.
	\label{eq.potts.relax.EL}
\end{align}
The gradient flow equation for  minimization problem (\ref{eq.potts.relax}) is: 
% to the following initial value problem (in the sense of gradient flow):
\begin{align}
	\begin{cases}
		\displaystyle\frac{\partial u}{\partial t} +F(f)-\lambda\varepsilon\nabla^2u +\frac{2\lambda}{\varepsilon}(2u^3-3u^2+u)=0 \mbox{ in }  \Omega\times (0,T],\\
		\displaystyle\frac{\partial u}{\partial \bn}=0 \mbox{ on } \partial \Omega,\\
		u(0)=G(f),
	\end{cases}
	\label{eq.potts.relax.gf}
\end{align}
for some initial condition $G(f)$, which is a function of $f$. Then solving (\ref{eq.potts.relax.EL}) is equivalent to finding the steady state solution of (\ref{eq.potts.relax.gf}). As was done in \cite{tai2023prep}, We add control variables to (\ref{eq.potts.relax.gf}) so that we can use these control variables to lead the final state $u(T)$ of the following equation to some desired targets: 
%directly and to better control the dynamics of $u$, we introduce control variable $W(t)$ and $b(t)$:
\begin{align}
	\begin{cases}
		\displaystyle \frac{\partial u}{\partial t} +F(f)-\lambda\varepsilon\nabla^2u +\frac{2\lambda}{\varepsilon}(2u^3-3u^2+u)+ W(t)*u+b(t)=0 \mbox{ in }  \Omega\times (0,T],\\
		\displaystyle \frac{\partial u}{\partial \bn}=0 \mbox{ on } \partial \Omega,\\
		u(0)=u_0.
	\end{cases}
	\label{eq.potts.relax.gf1}
\end{align}
Normally, the function $F(f)$ is a complicated function of $f$. We use a subnetwork to represent $F(\cdot)$.  As usual, here and later $*$ denote the spatial convolution operator. 

\subsection{An operator splitting method to solve (\ref{eq.potts.relax.gf1})}
We use the Lie scheme to solve (\ref{eq.potts.relax.gf1}). Still discretize the time variable as in Section {\ref{sec.splitting}.  Given $u^n$, we update $u^{n}\rightarrow u^{n+1/2} \rightarrow u^{n+1}$ in the following way:\\
\textbf{Substep 1}: Solve
\begin{align}
	\begin{cases}
		\displaystyle\frac{\partial v}{\partial t} +F(f)-\lambda\varepsilon\nabla^2v +W(f,t)*u+b(t)=0 \mbox{ in } \Omega\times(t^n,t^{n+1}],\\
		\displaystyle \frac{\partial v}{\partial \bn}=0 \mbox{ on } \partial \Omega,\\
		v(t^n)=u^n ,
	\end{cases}
	\label{eq.split.1}
\end{align}
and set $u^{n+1/2}=v(t^{n+1})$. \\
\textbf{Substep 2}: Solve 
\begin{align}
	\begin{cases}
		\displaystyle\frac{\partial v}{\partial t}+\frac{2\lambda}{\varepsilon}(2v^3-3v^2+v)=0 \mbox{ on } \Omega\times(t^{n},t^{n+1}],\\
		v(t^n)=u^{n+1/2},
	\end{cases}
	\label{eq.split.2}
\end{align}
and set $u^{n+1}=v(t^{n+1})$.
\subsection{Time discretization}
We use a one-step forward Euler scheme to time-discretize (\ref{eq.split.1}) and a one-step backward Euler scheme to time-discretize (\ref{eq.split.2}):
\begin{align}
	&\begin{cases}
		\displaystyle\frac{u^{n+1/2}-u^n}{\Delta t} +F(f)-\lambda\varepsilon\nabla^2u^n +W^n*u^n+b^n=0,
		\\
		\displaystyle\frac{\partial u^{n+1/2}}{\partial \bn}=0,
	\end{cases}\label{eq.split.1.dis}\\
	&	\displaystyle\frac{u^{n+1}-u^{n+1/2}}{\Delta t} +\frac{2\lambda}{\varepsilon}(2(u^{n+1})^3-3(u^{n+1})^2+u^{n+1})=0.
	\label{eq.split.2.dis}
\end{align}
In image processing, the periodic boundary condition is widely used. Here we replace the Neumann boundary condition in (\ref{eq.split.1.dis}) by the periodic one. In (\ref{eq.split.1.dis}), index $n$ is used at the superscript of $W^n$ and $b^n$ to differentiate control variables at different time.

In our algorithm, we choose $G(f)$ as a convolution layer of $f$, \red{i.e., a convolution of $f$ with a $3\times3$ kernel followed by a sigmoid function.} Suppose we are given a set of images $\{f_i\}_{i=1}^I$ and the corresponding segmentation masks $\{h_i\}_{i=1}^I$. \red{Note that both $F$ and $G$ are networks and contain learnable parameters.} Denote the set of control variables by $\theta_1=\{\{W^n, b^n,\}_{n=1}^N,\theta_F,\theta_G\}$, where $\theta_F,\theta_G$ denote the all network parameters in $F$ and $G$. Denote the procedure of numerically solving (\ref{eq.model1.control1}) with $N$ time steps and control variables $\theta_1$ by $$\cN_1(\theta_1;\cdot): f\rightarrow u^0\rightarrow u^1 \rightarrow \cdots \rightarrow u^N.$$ We will learn $\theta_1$ by solving
\begin{align}
	\min_{\theta_1} \frac{1}{I}\sum_{i=1}^I \ell(\cN_1(\theta_1;f_i),h_i),
	\label{eq.model2.learn}
\end{align}
where $\ell(\cdot,\cdot)$ is some loss function, such as the cross entropy.

\subsection{Connections to neural networks}
The building block for scheme (\ref{eq.split.1})--(\ref{eq.split.2}) consists of a linear step (\ref{eq.split.1.dis}) and a nonlinear step (\ref{eq.split.1.dis}). Thus a time stepping of $u^n$ corresponds to a layer of a network with (\ref{eq.split.2.dis}) being the activation function. Unlike the commonly used neural networks, the equivalent network of Model I has a heavy bias term $F(f)$ which is chosen as a subnetwork in this paper. The process (\ref{eq.model2.learn}) of learning $\theta_1$ is the same as training a network.

\subsection{Numerical experiments}

We demonstrate the effectiveness of Mode I. In our experiments, we choose $G(f)$ as a convolution layer of $f$. The functional $F(f)$ needs to be complicated enough to approximate the probability function in the Potts model. Here we set $F(f)$ as a UNet \cite{ronneberger2015u} subnetwork of $f$, \red{i.e., $F$ has the same architecture as a UNet, takes $f$ as the input and ouputs a matrix having the same size as $u$ with elements in $[0,1]$, see \cite{liu2023prep} for details.}  The $W^n$'s and $b^n$'s are convolution kernels and biases, respectively. \red{For parameters, we set $\Delta t=0.2,\lambda\varepsilon=1, \lambda/\varepsilon=15$.}

We use the MSRA10K dataset \cite{cheng2014global} which contains 10000 salient object images with manually annotated masks. We choose 2500 images for training and 400 images for testing. All images and masks are resized to $192\times 256$. 

We compare Model I with UNet \cite{ronneberger2015u}, UNet++ \cite{zhou2018unet++} and MANet \cite{fan2020ma}. All models are trained with 400 epochs. The comparison of accuracy and the dice score are shown in Table \ref{tab.model1.acc}. We observe that Model I has higher accuracy and dice score than other networks. Some segmentation results are presented in Figure \ref{fig.model1.results}. Model I successfully segments the targets from images with complicated structures.
\begin{table}
	\begin{tabular}{c|c|c|c|c}
		\hline
		& Model I & UNet & UNet++ & MANet\\
		\hline
		accuracy & {\bf 95.91\%} & 94.79\% & 95.66\% & 95.33\%\\
		\hline
		dice score & {\bf 0.9048} & 0.8776 & 0.8992 & 0.8936\\
		\hline
	\end{tabular}
\caption{Comparison of Model I with other networks in terms of accuracy and dice score.}
\label{tab.model1.acc}
\end{table}

\begin{figure}
	\begin{tabular}{cccc}
%		\hline
		Images & Given Mask & Mask by Model I & \makecell{Segmented Image\\by Model I}\\
%		\hline
		\includegraphics[width=0.18\textwidth]{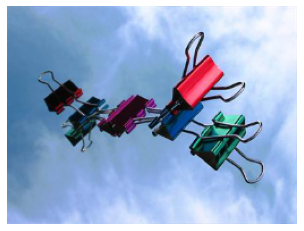} &
		\includegraphics[width=0.18\textwidth]{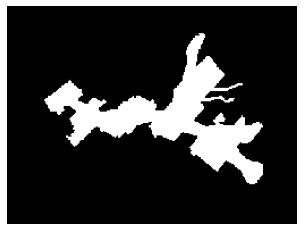} &
		\includegraphics[width=0.18\textwidth]{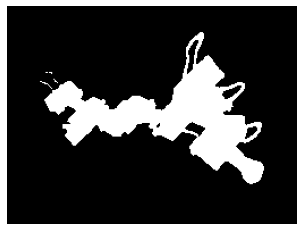} &
		\includegraphics[width=0.18\textwidth]{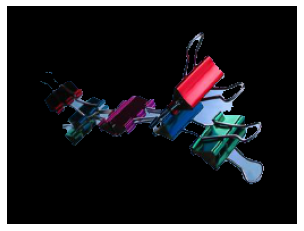}\\
		\includegraphics[width=0.18\textwidth]{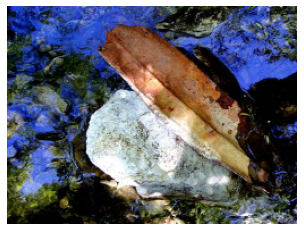} &
		\includegraphics[width=0.18\textwidth]{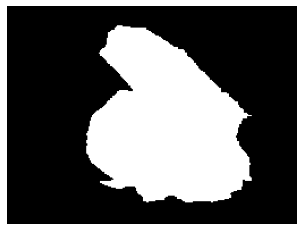} &
		\includegraphics[width=0.18\textwidth]{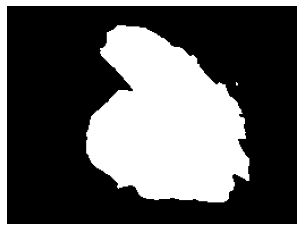} &
		\includegraphics[width=0.18\textwidth]{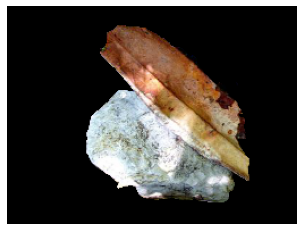}\\
		\includegraphics[width=0.18\textwidth]{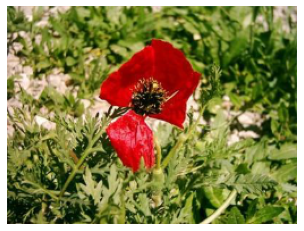} &
		\includegraphics[width=0.18\textwidth]{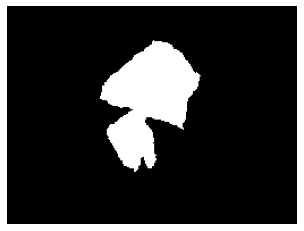} &
		\includegraphics[width=0.18\textwidth]{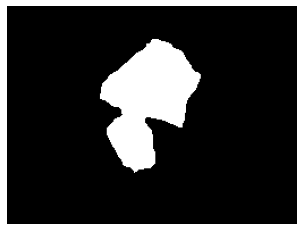} &
		\includegraphics[width=0.18\textwidth]{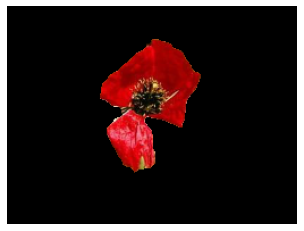}\\
	\end{tabular}
\caption{Some results by Model I.}
\label{fig.model1.results}
\end{figure}
\section{Operator-splitting method inspired networks for image segmentation: Model II}
\label{sec.model2}
As in \cite{tai2023prep},  
our second model incorporates (\ref{eq.potts}) and the threshold dynamics ideas in approximating the perimeter of a region. 
\subsection{Model formulation}
\red{In (\ref{eq.potts}), when $v$ is binary, $\int_{\Omega}|\nabla v| d\bx$ gives the perimeter of $\Omega_1$, the support of $v$. Replacing $\int_{\Omega}|\nabla v| d\bx$ by $\Per(\Omega_1)$ in (\ref{eq.potts}), we get}
%In the first model, we use the exact perimeter term in (\ref{eq.potts}) to get
\begin{align}
	\min_{v\in\{0,1\}}\int_{\Omega} F(f)vd\bx +\lambda \Per(\Omega_1).
	\label{eq.model1.1}
\end{align}
Then we use a threshold dynamics idea to approximate the perimeter term:	
\begin{align}
	 	\Per(\Omega_1)  \red{\approx}  	\sqrt{\frac{\pi}{\red{\delta}}} \int_{\Omega} v (\bx)(G_{\red{\delta}}*(1-v )(\bx)d\bx,
	\label{eq.per.dynamic}
\end{align}
where $G_{\red{\delta}}$ is the \red{two-dimensional Gaussian filter (the covariance matrix of $\bx$ being an identity matrix)} 
\begin{align}
	G_{\red{\delta}}(\bx)=\frac{1}{2\pi\red{\delta}^2} \exp\left(-\frac{\|\bx\|^2}{2\red{\delta}^2}\right).
\end{align}
It is shown in \cite{pallara2007short} that the approximation in (\ref{eq.per.dynamic}) $\Gamma$--converges to $\Per(\Omega_1) $ as $\red{\delta}\rightarrow 0$.
Replacing $\Per(\Omega_1)$ by the approximation given in (\ref{eq.per.dynamic}),
the functional we are minimizing becomes 
\begin{align}
	\min_{v\in\{0,1\}}\int_{\Omega} F(f)vd\bx +\lambda \int_{\Omega} v (\bx)(G_{\red{\delta}}*(1-v )(\bx)d\bx.
	\label{eq.model1.2}
\end{align}
Model (\ref{eq.model1.2}) requires $v$ to be binary. We relax this constraint to $v\in [0,1]$ and consider the following functional instead
\begin{align}
	\min_{v\in [0,1]}\int_{\Omega} F(f)vd\bx + \varepsilon\int_{\Omega} v  \ln v + (1-v)\ln(1-v) d\bx+\lambda  \int_{\Omega} v (\bx)(G_{\red{\delta}}*(1-v )(\bx)d\bx
	\label{eq.model1.3}
\end{align}
for some small $\varepsilon>0$. Such an approximate is a smoothed version of the binary constraint, and it converges to the original problem (\ref{eq.model1.2}) as $\varepsilon\rightarrow 0$, c.f.  \cite{liu2022deep}.

If $u$ is a minimizer of (\ref{eq.model1.3}), it satisfies the optimality condition
$$ \red{\varepsilon\left( \ln \frac{u}{1-u}\right)}   + \lambda G_{\red{\delta}} *(1-2u ) + F(f)  =  0. $$
The gradient flow equation for this problem is: 
\begin{align}
	\begin{cases}
		\displaystyle\frac {\partial u}{\partial t}   + \varepsilon\left( \ln \frac{u}{1-u}\right)  + \lambda G_{\red{\delta}} *(1-2u) + F(f)  = 0  ,\\  
		u(0) = u_0 . 
	\end{cases}
\label{eq.model1.ivp}
\end{align}
We then introduce control variables to (\ref{eq.model1.ivp}) and solve
\begin{align}
	\begin{cases}
		\displaystyle\frac {\partial u}{\partial t}   + \varepsilon\left( \ln \frac{u}{1-u}\right)  + \lambda G_{\red{\delta}} *(1-2u) + F(f)+A(t)*u+g(t)  = 0  ,\\  
		u(0) = u_0=G(f)
	\end{cases}
	\label{eq.model1.control}
\end{align}
for some initial condition $G(f)$ which is a function applied on $f$. Here we used a different notation from Section \ref{sec.model1} to differentiate the different control variables. 

Unlike Model I which treats $F(f)$ and $b(t)$ as two functions, we use a different strategy to treat Model II. Since the term $F(f)$ and $g$ are just constant terms with respect to $u$, we can combine them and denote them by $g$ which depends on $f$. The new problem is written as
\begin{align}
	\begin{cases}
		\displaystyle\frac {\partial u}{\partial t}   + \varepsilon\left( \ln \frac{u}{1-u}\right) + \lambda G_{\red{\delta}} *(1-2u) + A(t)*u+g(f,t)  = 0  ,\\  
		u(0) = u_0=G(f) . 
	\end{cases}
	\label{eq.model1.control1}
\end{align}

\subsection{An operator splitting method to solve (\ref{eq.model1.control1})}
We decompose the operator $A$ and the function $g$ as a sum of $K$ terms for some positive integer $K$:
\begin{align}
	A=\sum_{k=1}^K A_k, \quad g=\sum_{k=1}^K g_k
\end{align}
for some operators $A_k$'s and functions $g_k$'s. We then use a Lie scheme to solve (\ref{eq.model1.control1}). Given $u^n$, we update $u^n\rightarrow \cdots \rightarrow u^{n+\frac{k}{K}} \rightarrow \cdots \rightarrow u^{n+1}$ via $K$ substeps:

\noindent For $k=1,...,K-1$, solve
\begin{align}
	\begin{cases}
		\displaystyle\frac{\partial v}{\partial t}   + \varepsilon\left( \ln \frac{v}{1-v}\right)   + A_k(t)*v+g_k(f,t)  = 0 \mbox{ in } \Omega\times (t^n,t^{n+1}],\\
		\displaystyle v(t^n)=u^{n+\frac{k-1}{K}},
		\label{eq.model1.stepk}
	\end{cases}
\end{align}
and set $u^{n+\frac{k}{K}}=v(t^{n+1})$.

\noindent For $k=K$, solve
\begin{align}
	\begin{cases}
		\displaystyle\frac{\partial v}{\partial t}   + \varepsilon\left( \ln \frac{v}{1-v}\right) + \lambda G_{\red{\delta}} *(1-2v)  +A_K(t)*v+g_K(f,t)  = 0 \mbox{ in } \Omega\times (t^n,t^{n+1}],\\
		\displaystyle v(t^n)=u^{n+\frac{K-1}{K}},
		\label{eq.model1.stepK}
	\end{cases}
\end{align}
and set $u^{n+1}=v(t^{n+1})$. 

For (\ref{eq.model1.stepk}), we further apply a Lie scheme to decompose it into two substeps:

\noindent{\bf Substep 1:}
Solve
\begin{align}
	\begin{cases}
		\displaystyle \frac{\partial v}{\partial t}   +  A_k(t)*v+g_k(f,t)  = 0 \mbox{ in } \Omega\times (t^n,t^{n+1}],\\
		v(t^n)=u^{n+\frac{k-1}{K}},
		\label{eq.model1.stepk.1}
	\end{cases}
\end{align}
and set $\bar{u}^{n+\frac{k}{K}}=v(t^{n+1})$.

\noindent{\bf Substep 2:}
Solve
\begin{align}
	\begin{cases}
		\displaystyle\frac{\partial v}{\partial t}   + \varepsilon\left( \ln \frac{v}{1-v}\right)  = 0 \mbox{ in } \Omega\times (t^n,t^{n+1}],\\
		v(t^n)=\bar{u}^{n+\frac{k}{K}},
		\label{eq.model1.stepk.2}
	\end{cases}
\end{align}
and set $u^{n+\frac{k}{K}}=v(t^{n+1})$.

Similarly, we solve (\ref{eq.model1.stepK}) by the following two substeps:

\noindent{\bf Substep 1:}
Solve
\begin{align}
	\begin{cases}
		\displaystyle \frac{\partial v}{\partial t}   +  A_K(t)*v+g_K(f,t)  = 0 \mbox{ in } \Omega\times (t^n,t^{n+1}],\\
		v(t^n)=u^{n+\frac{K-1}{K}},
		\label{eq.model1.stepK.1}
	\end{cases}
\end{align}
and set $\bar{u}^{n+1}=v(t^{n+1})$.

\noindent{\bf Substep 2:}
Solve
\begin{align}
	\begin{cases}
		\displaystyle \frac{\partial v}{\partial t}   + \varepsilon\left( \ln \frac{v}{1-v}\right) + \lambda G_{\red{\delta}} *(1-2u)  = 0 \mbox{ in } \Omega\times (t^n,t^{n+1}],\\
		v(t^n)=\bar{u}^{n+1},
		\label{eq.model1.stepK.2}
	\end{cases}
\end{align}
and set $u^{n+1}=v(t^{n+1})$. 

\subsection{Time discretization}
Problem (\ref{eq.model1.stepk.1})--(\ref{eq.model1.stepK.2}) are only semi-constructive as we still need to solve these initial value problems. In our algorithm, we time-discretize (\ref{eq.model1.stepk.1}) and (\ref{eq.model1.stepK.1}) by the forward Euler method and discretize (\ref{eq.model1.stepk.2}) and (\ref{eq.model1.stepK.2}) by the backward Euler method. The discretized scheme is given as follows: For (\ref{eq.model1.stepk}), we use the following time discretization to solve it
\begin{align}
	&\frac{\bar{u}^{n+\frac{k}{K}}-u^{n+\frac{k-1}{K}}}{\Delta t}+ A^n_k*u^{n+\frac{k-1}{K}} +g^n_k=0,
	\label{eq.model1.stepk.1.dis}\\
	&\frac{u^{n+\frac{k}{K}}-\bar{u}^{n+\frac{k}{K}}}{\Delta t}+\red{\varepsilon}\left( \ln \frac{u^{n+\frac{k}{K}}}{1-u^{n+\frac{k}{K}}}\right)=0.
	\label{eq.model1.stepk.2.dis}
\end{align}
For (\ref{eq.model1.stepK}), we use the following time discretization to solve it
\begin{align}
	&\frac{\bar{u}^{n+1}-u^{n+\frac{K-1}{K}}}{\Delta t}+ A^n_K*u^{n+\frac{K-1}{K}} +g^n_K=0,
	\label{eq.model1.stepK.1.dis}\\
	&\frac{u^{n+1}-\bar{u}^{n+1}}{\Delta t}+\red{\varepsilon}\left( \ln \frac{u^{n+1}}{1-u^{n+1}}\right) + \red{\lambda G_\delta *(1-2u^{n+1})}=0.
	\label{eq.model1.stepK.2.dis}
\end{align}

In our algorithm, we choose $G(f)$ as a convolution layer of $f$, \red{i.e., a convolution of $f$ with a $3\times 3$ kernel followed by a sigmoid function. We will also choose $g^n$'s as networks.}  Suppose we are given a set of images $\{f_i\}_{i=1}^I$ and the corresponding segmentation masks $\{h_i\}_{i=1}^I$. Denote the set of control variables by $\theta_2=\{A^n,\theta_{g^n},\theta_{G}\}_{n=1}^N$, where $\theta_{g^n}$ and $\theta_G$ denote the parameters in $g^n$ and $G$, respectively. Denote the procedure of numerically solving (\ref{eq.model1.control1}) with $N$ time steps and control variables $\theta_2$ by $$\cN_2(\theta_2;\cdot): f\rightarrow u^0\rightarrow u^1\rightarrow \cdots \rightarrow u^N.$$ We will learn $\theta_2$ by solving 
\begin{align}
	\min_{\theta_2} \frac{1}{I}\sum_{i=1}^I \ell(\cN_2(\theta_2;f_i),h_i),
	\label{eq.model1.learn}
\end{align}
where $\ell(\cdot,\cdot)$ is some loss function, such as the cross entropy.
\subsection{Connections to neural networks}
The building block  of scheme (\ref{eq.model1.stepk})--(\ref{eq.model1.stepK}) is  (\ref{eq.model1.stepk.1.dis})--(\ref{eq.model1.stepK.2.dis}). Note that the first substep (\ref{eq.model1.stepk.1.dis}) (resp. (\ref{eq.model1.stepK.1.dis})) is a linear step in $\bar{u}^{n+\frac{k}{K}}$ (resp. $\bar{u}^{n+1}$). The second step (\ref{eq.model1.stepk.2.dis}) (resp. (\ref{eq.model1.stepK.2.dis})) is a nonlinear step. Thus this procedure is the same as a layer of a neural network, which consists a linear step and a nonlinear step (activation step). The numerical scheme for (\ref{eq.model1.stepk})--(\ref{eq.model1.stepK}) that maps $f\rightarrow u^0\rightarrow u^1 \rightarrow \cdots \rightarrow u^N$ is a neural network with $KN$ layers and activation functions specified in (\ref{eq.model1.stepk.2.dis}) and (\ref{eq.model1.stepK.2.dis}). The process (\ref{eq.model1.learn}) of learning $\theta_2$ is the same as training a network.

\subsection{Numerical experiments}

In this section, we demonstrate the robustness of Model II against noise. We show that one trained Model II can provide good segmentation results on images with various levels of noise. In our experiments, we choose $G(f)$ as a convolution layer of $f$. We set $A^n_k$'s as learnable convolution kernels at different scales that extract various image features. We set $g_k^n$'s as the convolution layers that are applied to $f$. \red{For parameters, we use $\Delta t=0.5,\varepsilon=2, \lambda=80$.}

We use the MSRA10K data again while resizing all images to a size of $192\times 256$. In our training, we train our model on images with noise standard deviation (SD) 1.0. This noise is big. Many existing algorithms cannot handle so big amount of noise. Some segmentation results are presented in Figure \ref{fig.model2.results}. For various levels of noise (even with very large noise), the trained model segments the target well. Refer to \cite{tai2023prep} for more experiments and explanations. 

\begin{figure}
	\begin{tabular}{ccccc}
		\hline\hline
		\makecell[c]{Clean Image\\ and Mask} &\includegraphics[align=c,width=0.18\textwidth]{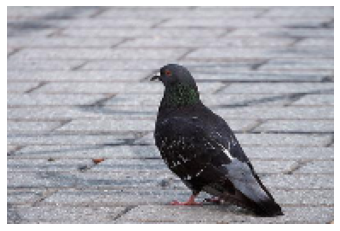} & 
		\includegraphics[align=c,width=0.18\textwidth]{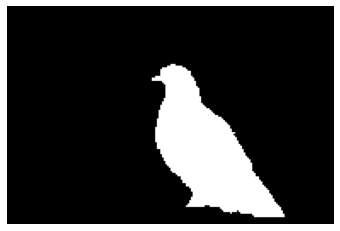}\\
		&SD=0 & SD=0.2 & SD=0.4 & SD=0.7\\
		Given Image & \includegraphics[align=c,width=0.18\textwidth]{figures/MA-all-SD00-14-5-img} &
		\includegraphics[align=c,width=0.18\textwidth]{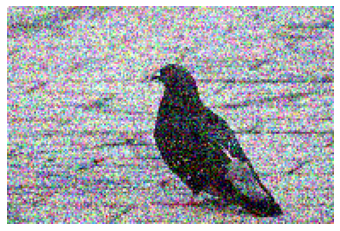} &
		\includegraphics[align=c,width=0.18\textwidth]{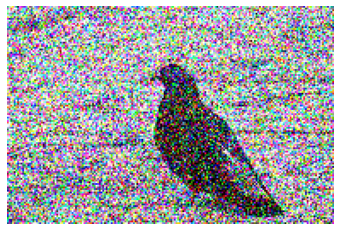} &
		\includegraphics[align=c,width=0.18\textwidth]{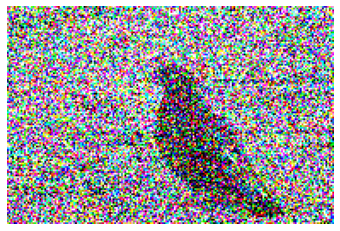} \\
		\makecell{Results by\\Model II} & 
		\includegraphics[align=c,width=0.18\textwidth]{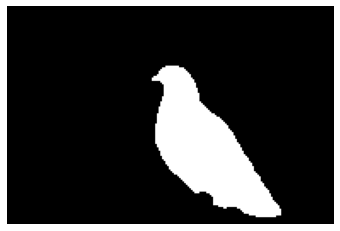} &
		\includegraphics[align=c,width=0.18\textwidth]{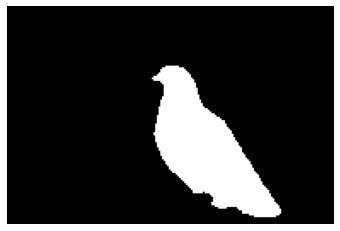} &
		\includegraphics[align=c,width=0.18\textwidth]{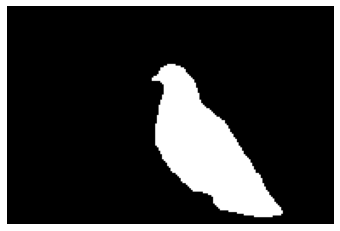} &
		\includegraphics[align=c,width=0.18\textwidth]{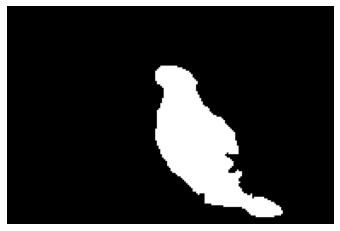} \\
		\makecell{Segmented\\Clean Images} & 
		\includegraphics[align=c,width=0.18\textwidth]{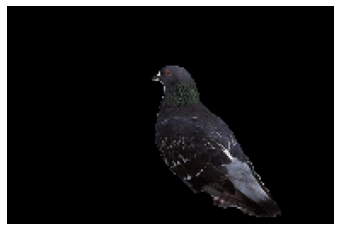} &
		\includegraphics[align=c,width=0.18\textwidth]{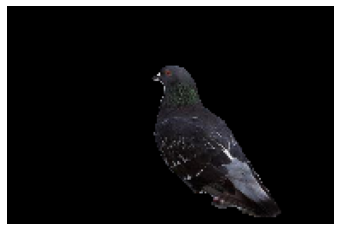} &
		\includegraphics[align=c,width=0.18\textwidth]{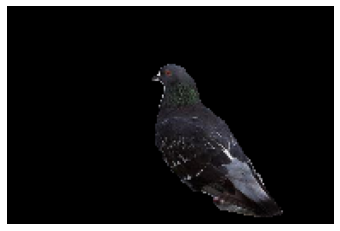} &
		\includegraphics[align=c,width=0.18\textwidth]{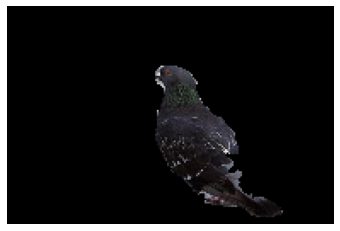}\\
		\hline \hline
		\makecell[c]{Clean Image\\ and Mask} &\includegraphics[align=c,width=0.18\textwidth]{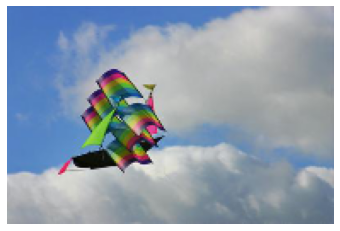} & 
		\includegraphics[align=c,width=0.18\textwidth]{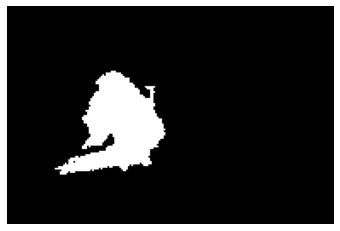}\\
		&SD=0 & SD=0.2 & SD=0.4 & SD=0.7\\
		Given Image & \includegraphics[align=c,width=0.18\textwidth]{figures/MA-all-SD00-40-3-img} &
		\includegraphics[align=c,width=0.18\textwidth]{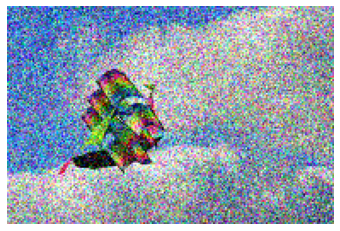} &
		\includegraphics[align=c,width=0.18\textwidth]{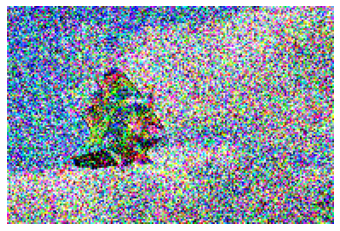} &
		\includegraphics[align=c,width=0.18\textwidth]{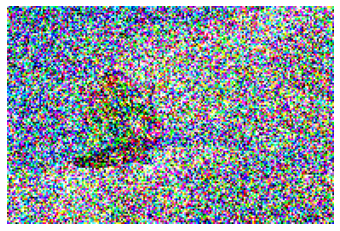} \\
		\makecell{Results by\\Model II} & 
		\includegraphics[align=c,width=0.18\textwidth]{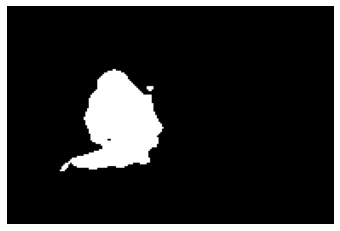} &
		\includegraphics[align=c,width=0.18\textwidth]{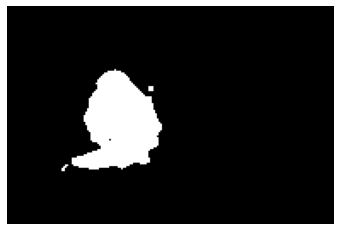} &
		\includegraphics[align=c,width=0.18\textwidth]{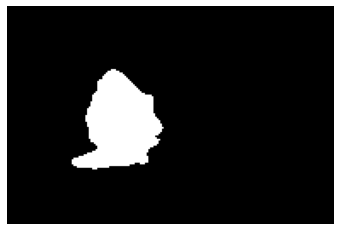} &
		\includegraphics[align=c,width=0.18\textwidth]{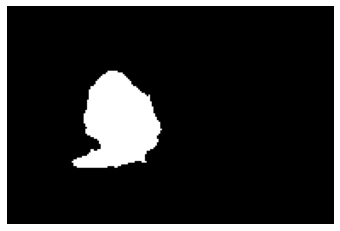} \\
		\makecell{Segmented\\Clean Images} & 
		\includegraphics[align=c,width=0.18\textwidth]{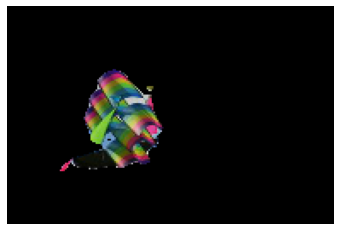} &
		\includegraphics[align=c,width=0.18\textwidth]{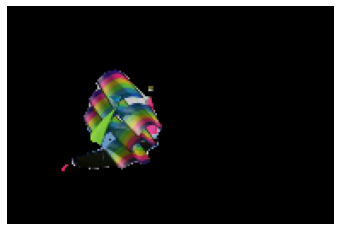} &
		\includegraphics[align=c,width=0.18\textwidth]{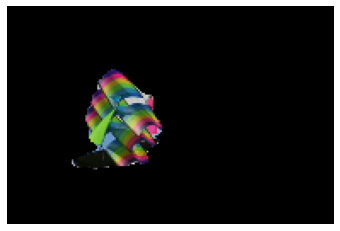} &
		\includegraphics[align=c,width=0.18\textwidth]{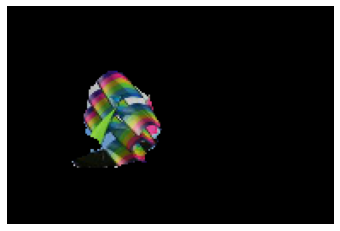}\\
		\hline\hline
	\end{tabular}
\caption{Results by Model II.}
\label{fig.model2.results}
\end{figure}

\section{Conclusion}
\label{sec.conclusion}

In this paper, the relations between deep neural networks and operator-splitting methods are discussed, and two new networks inspired by the operator-splitting method are proposed for image segmentation. The two proposed networks are derived from the Potts model, with certain terms having physical meanings as regularizers. Essentially, the two networks are operator-splitting algorithms solving the Potts model. The effectiveness of the proposed networks is demonstrated by numerical experiments.

\bibliographystyle{abbrv}
\bibliography{ref}
\end{document}